\pdfoutput=1

\documentclass[11pt]{article}

\usepackage[final]{acl}

\usepackage{times}
\usepackage{latexsym}
\usepackage{ulem}
\usepackage{array}
\usepackage{ragged2e}
\usepackage{amsmath}
\usepackage{subcaption}

\usepackage{graphicx} 
\usepackage{booktabs}
\usepackage{xspace}

\definecolor{clusterblue}{HTML}{4EA3D6}
\definecolor{clustergreen}{HTML}{9DD7BC}
\definecolor{clusteryellow}{HTML}{F1D795}
\definecolor{clusterred}{HTML}{F1A077}

\newcommand{\ColorfulFrameworkName}{
\textsc{\textcolor{clusterblue}{S}\textcolor{clustergreen}{k}\textcolor{clusteryellow}{i}\textcolor{clusterred}{l}\textcolor{clusterblue}{l}\textcolor{clustergreen}{V}\textcolor{clusteryellow}{e}\textcolor{clusterred}{r}\textcolor{clusterblue}{s}\textcolor{clustergreen}{e}}
}

\newcommand{\FrameworkName}{\textsc{SkillVerse}\xspace}
\usepackage[T1]{fontenc}

\usepackage[utf8]{inputenc}

\usepackage{microtype}

\usepackage{inconsolata}
\usepackage{colortbl}
\usepackage{float}

\usepackage{graphicx}

\title{\ColorfulFrameworkName: Assessing and Enhancing LLMs with Tree Evaluation}

\author{Yufei Tian~\thanks{\: Work done when the first author was interning at Google.} $^{1}$
\quad Jiao Sun$^{2}$ \quad Nanyun Peng$^{1}$ \quad  Zizhao Zhang$^{3}$\\[7pt]
         $^1$University of California, Los Angeles,\\
         $^2$Google DeepMind, $^3$Google Cloud AI \\[5pt]
         {
         \texttt{yufeit@cs.ucla.edu} 
         }
         }

\begin{document}
\maketitle
\begin{abstract}

As language models evolve to tackle complex and multifaceted tasks, their evaluation must adapt to capture this intricacy. A granular, skill-specific understanding of model capabilities can empower researchers to make informed model development plans.
In this paper, we introduce \FrameworkName{}, an unsupervised tree-structured diagnosis framework for understanding model proficiency in specific abilities. With LLM as a judge, \FrameworkName{} first critiques the model responses, and then organizes them into a hierarchical structure termed dendrogram. 
Given proficiency at arbitrary levels of granularity, \FrameworkName{} is flexible to produce insights of behaviors of modern large models. We also demonstrate its efficacy in two downstream tasks: 1) improving model in-context learning by 25\% using a tree-search algorithm to select more informative few shots, and 2) accurately predicting new model weaknesses with a 55\% success rate, 22\% higher than the baseline.


\end{abstract}

\section{Introduction}

In recent years, leaderboard and benchmark results such as ChatbotArena \cite{chiang2024chatbot} and MMLU \cite{hendrycks2020measuring} have become the dominant practice for evaluating the potency of language models. 
While these results provide a high-level snapshot of a model's rank, their limited interpretability makes it difficult to identify subtle behavioral traits and derive actionable insights \cite{murahari2023qualeval, moayeri2024unearthing}. 

The limited interpretability of the current evaluation paradigm makes it hard to compare the relative strengths and weaknesses of different models. For instance, does a higher-ranked model consistently outperform lower-ranked counterparts across the entire benchmark? Do comparable scores transfer to equivalent model performance on all subdomains? Addressing such questions typically requires manual inspections, which is both time-consuming and costly. These challenges highlight the need for automatic, granular analyses that provide valuable insights to enrich our understanding of model behavior, which paves the road for targeted model improvement of specific capabilities.

\begin{figure}[t!]
    \centering
    \includegraphics[width=0.95\linewidth]{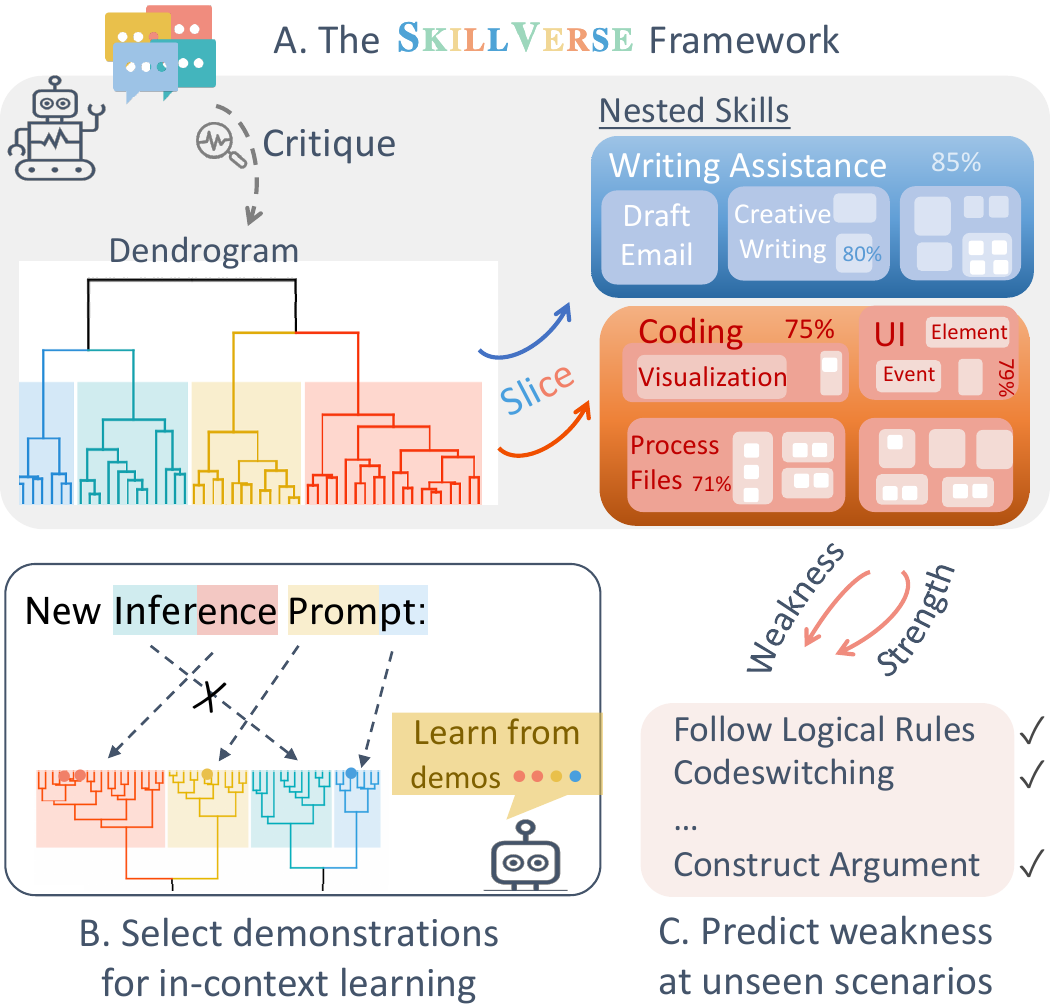}
    \vspace{-2mm}
    \caption{\textbf{Up:} The input-output flow of \FrameworkName{}. Skill-specific critiques are extracted, structured into a dendrogram, and sliced at varying granularities to reveal nested clusters of skills and model proficiency. 
    \textbf{Bottom:} Versatile applications of \FrameworkName{}, from selecting informative few-shot demonstrations to uncovering hidden model weaknesses.
}
    \label{fig:teaser}
    \vspace{-1em}
\end{figure}

\begin{figure*}[t!]
    \centering
    \includegraphics[width=0.99
\linewidth]{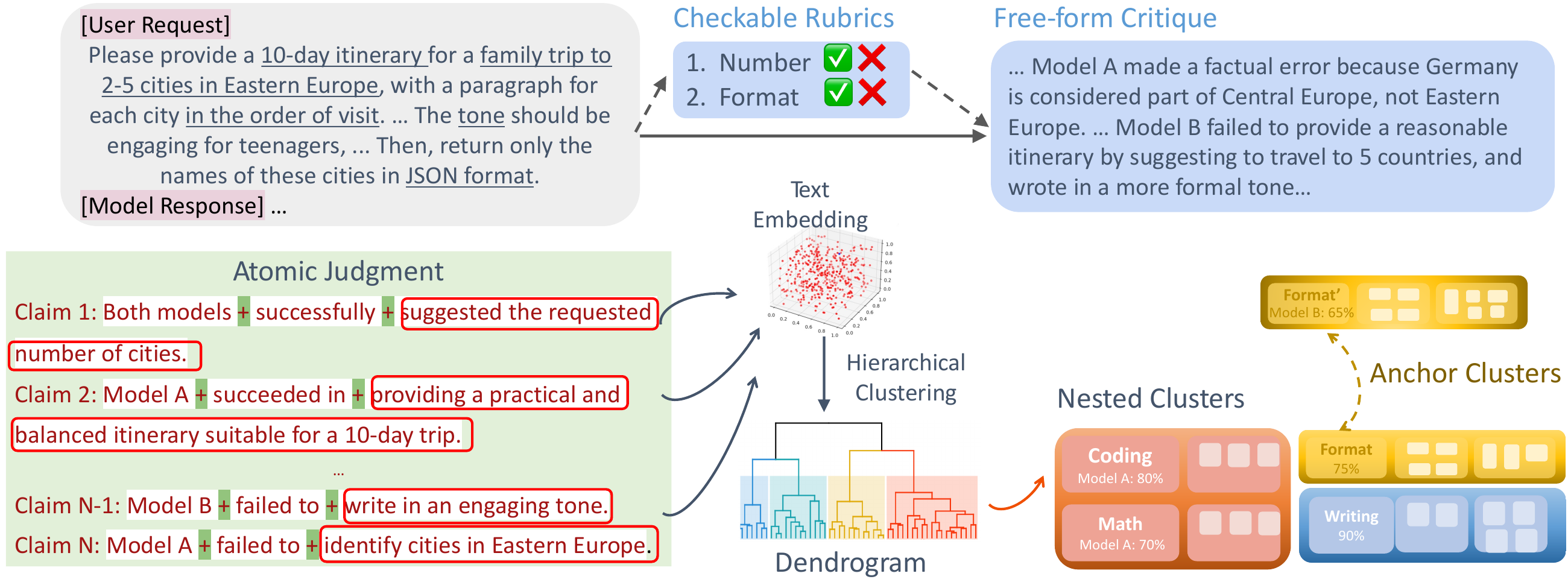}
    \vspace{-3mm}
    \caption{The overall framework of \FrameworkName{}. A set of critiques on model responses is parsed into atomic judgments and organized using bottom-up clustering into a dendrogram, which is then unfolded at varying levels of granularity to form nested clusters, allowing for detailed analysis of model proficiency. Thanks to the hierarchical structure, this novel pipeline is highly flexible in interpreting model capabilities. 
    }
    \vspace{-1em}
    \label{fig:overview}
   
\end{figure*}

Lately, LLM-based evaluations \cite{alpaca_eval,zheng2023judging}, where LLMs critique and judge model responses, have emerged as a scalable approach to approximate human preferences \cite{wang2023shepherd, yuan2024self}. These methods enable detailed analysis with rich contextual feedback, forming the foundation of our diagnosis framework, \FrameworkName{}. Orthogonal to developing more reliable auto-raters, this paper contributes to \textbf{structuring contextual feedback} to generate actionable insights for model model evaluation, comparison, debugging, and improvement.


As shown in Figure \ref{fig:teaser} and Figure \ref{fig:overview}, \FrameworkName{} generates personalized insights through a tree-structured model assessment, tailored to the level of detail preferred by human scientists. To quantify and extract actionable insights from critiques to diverse real-world data, we introduce \textit{atomic judgment}: an assessment of an indivisible aspect of model capability. Next, we conduct agglomerative clustering on these atomic judgments based on their semantic distance, resulting in a dendrogram. This tree can be chopped at different levels, resulting into clusters of varying sizes or granularities. Each cluster represents a specific skill, for which we calculate success rates to evaluate model performance ($\S$\ref{sec:skillverse}).

\FrameworkName{} produces insights of current model behaviors, such as Gemini, Claude, and GPT-4 ($\S$\ref{sec:result}). 
For instance, on Arena-Hard benchmark \cite{li2024crowdsourced} where Claude-3.5-Sonnet \cite{anthropic2024claude35sonnet} ranks 2nd and Gemini-1.5-Pro \cite{gemini2024} ranks 6th,\footnote{Under \texttt{Arena-Hard-Auto} on \url{https://lmarena.ai}, as of Nov 18th, 2024.} \FrameworkName{} finds out Gemini needs improvement in debugging, writing command lines, and business analysis. On the other hand, the higher-ranked  Claude, falls short in providing analogical examples, debate, and evaluate arguments. Notably, by comparing different sized models within the same family, we also identify instances of inverse scaling, where larger models underperform smaller ones due to strong parameterized knowledge \cite{mckenzieinverse}. Such tasks include handling word inclusion/exclusion, word counts, and adherence to specific formats.

To validate \FrameworkName{}'s ability to identify true model errors and its potential values for model improvement, we design a few extended improvement explorations that lead to promising gains. 
In $\S$\ref{sec:ICL}, we show that the dendrogram enhances in-context learning by enabling a tree search algorithm that adaptively selects challenging and relevant examples as contrastive few-shot demonstrations. This approach achieves a 25\% relative improvement over the standard contrastive in-context learning method (C-ICL, \citet{Yan2021SemanticsGuidedCN}). In $\S$\ref{sec:extrapolate}, we demonstrate that a strong reasoner such as GPT-4o can digest the model proficiency report generated from \FrameworkName{}  to predict weaknesses in unseen scenarios. For example, the model proficiency on ten hypothesized tasks is only 55\%, 22\% lower than uninformed predictions by the same reasoner. 

We showcase that \FrameworkName{} can serve as a powerful tool for providing fine-grained interpretation of model behaviors and developing targeted improvement of discovered model deficiencies during inference. Future research could also leverage the actionable feedback derived from \FrameworkName{} to a wide range tasks: such as model routing, curatiing targeted training data for specific subdomains where the current model underperforms, and etc.



\begin{figure*}[t!]
    \centering
    \includegraphics[width=0.99
\linewidth]{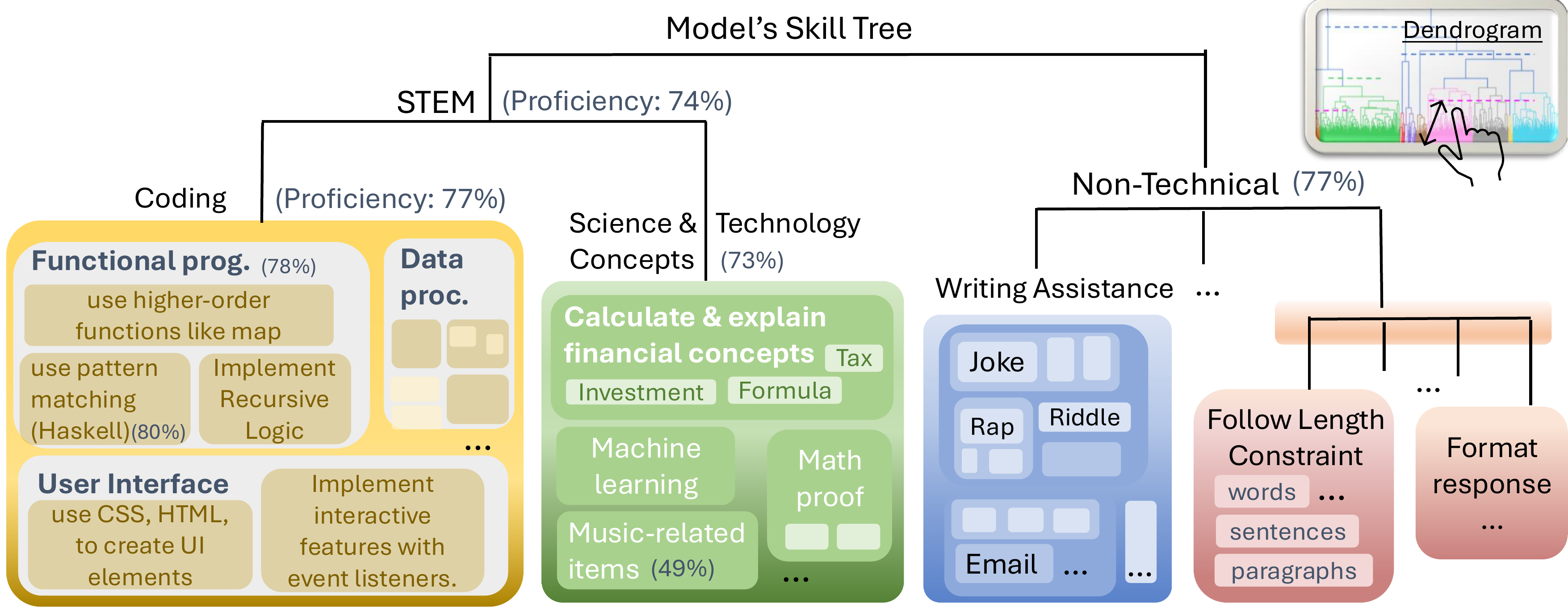}
    \vspace{-2mm}
    \caption{\textbf{Upper right}: A dendrogram produced by \FrameworkName{} on a combination of two datasets: ChatbotArena \cite{chiang2024chatbot} and IF Eval \cite{zhou2023instruction}. \textbf{Main figure}:  Multiple layers of nested clusters that represent model proficiencies from coarse-grained to fine-grained, by horizontally slicing the dendrogram at different levels. For each group, we can then calculate skill-level model proficiency based on the atomic judgments, as demonstrated in the parenthesis. An LLM summarizes all members in the same cluster and generate the skill-level description.
}\vspace{-1em}
    \label{fig:unfold}

\end{figure*}


\section{\ColorfulFrameworkName: Diagnosis Framework} \label{sec:skillverse}

\subsection{Overview}
Figure \ref{fig:overview} presents the overall framework of \FrameworkName{}. Starting with a large set of user prompts and model responses, we collect critiques that evaluate model responses in detail ($\S$\ref{subsec:collect_critique}). These critiques are then parsed into atomic judgments, enabling efficient organization and large-scale quantification. Using bottom-up clustering algorithms, the atomic judgments are structured into a dendrogram. To interpret the results, the dendrogram is unfolded at varying levels of granularity based on the detail preferred by engineers and researchers ($\S$\ref{subsec:structure_critique} and Figure \ref{fig:unfold}). Finally, clusters derived independently are anchored to support multi-party analyses ($\S$\ref{subsec:anchor}).  

\subsection{Collecting Accurate Critiques}\label{subsec:collect_critique}
A straightforward way to collect critiques is adopting language models off-the-shelf or finetune them on domain specific data---a reliable approach for evaluating content relevance, style. etc. For a comprehensive representation of model capabilities, our critique model takes in both positive and negative samples and is capable of identifying both the weaknesses and strengths of a model's response. 

However, recent studies \cite{murugadoss2024evaluating, sonllm, jing2024scale} highlight limitations of LLMs-as-a-Judge in domains like factual verification, format checking, and calculations. Fortunately, aspects like format and calculation are programmatically checkable, eliminating the need to rely solely on language models for feedback.

\paragraph{Checkable Rubrics} To enhance accuracy, we first identify the checkable components of a user request and leverage programs to evaluate these metrics. Previous work on instruction following \cite{zhou2023instruction} identified and open-sourced 25 types of verifiable instructions for writing tasks (\textit{e.g.}, multiple sections, forbidden words, numbers). We built upon their efforts to create a similar database of verifiers for these checkable subtasks. In practice, we first identify whether any part of a given instruction can be mapped to our database of checkable subtasks. If a match is found, a tuple of `{target checkable task, user request, model response}` is passed to our program, which produces a verified result. These results are then provided as input to the critique model, alongside the original user prompt and model response, resulting in a more robust evaluation.

\subsection{Structuring Diverse Critiques}\label{subsec:structure_critique}

\paragraph{Converting to Atomic Judgments}
To efficiently organize thousands of free-form critiques, we introduce the concept of atomic judgments, which serve as act as the building blocks for systematically organizing these critiques. An atomic claim is a statement that addresses a single, non-decomposable aspect of model ability (\textit{e.g.}, ``\textit{Model A + failed to + identify cities in Easter Europe.}'' as is illustrated in Figure 2). We enforce all atomic judgments to follow a strict syntax with three components: Subject  (\textit{i.e.}, the model name) + Verb (\textit{i.e.}, succeed, partially succeed, or fail) + Object (\textit{i.e.}, a specific task), which provides the necessary certainty and precision to quantify large volumes of critiques and \uline{calculate model proficiency}.

As a preparation step for clustering, we leverage Google's Text Embedding API \cite{GoogleTextEmbeddingAPI} to vectorize these atomic judgments.
Since the both the subject and verb are deterministic, we focus exclusively on embedding and clustering the third component, the specific task. 

\paragraph{Hierarchical Clustering}
We perform agglomerative (\textit{i.e,} bottom-up) clustering on the atomic judgments based on their semantic distance. The algorithm begins by treating each claim as an individual cluster. It then identifies the two closest clusters and merges them. This unsupervised process is repeated until all the clusters are merged into a single one, resulting in a tree of nested clusters, also known as a dendrogram.

\paragraph{Interpreting the Dendrogram}
To analyze model behavior, we horizontally slice the dendrogram at a preferred level. Next, to obtain a description for each resulting cluster, we prompt an LLM to summarize the group members. 

Figure \ref{fig:unfold} illustrates a dendrogram produced in one of our experiments on the ChatbotArena \cite{chiang2024chatbot} and the IF Eval dataset \cite{zhou2023instruction}. This hierarchy captures relations among all data points. A horizontal cut at the highest level yields two primary branches: a left, technical branch and a right, non-STEM branch. Further slicing these two branches reveals subclusters: 1) \textit{coding}, 2) \textit{calculating formulas and explaining STEM concepts} for the left and 1) \textit{format output}, 2) \textit{providing helpful information}, 3) \textit{writing assistance and other content creation} for the right one. Each cluster is further nested into smaller subclusters (refer to Figure \ref{fig:unfold} for details). Model proficiency can be computed by calculating the ratio of positive atomic judgments within these clusters.


\subsection{Anchoring Clusters}\label{subsec:anchor}
 
As different models often generate varying responses, the corresponding dendrograms produced by clustering may differ slightly. Additionally, re-running the clustering process with all existing model responses every time a new model is added is inefficient. To support multi-party analyses, it is essential to merge or anchor clusters derived independently and ensure consistency across models.

Our algorithm merge two clusters only if two key conditions are satisfied. First, the centroids of the clusters must be close to each other in the feature space. Let $\mu_i$ and $\mu_j$ denote the centroids of clusters $C_i$ and $C_j$. The clusters can be merged if their similarity exceeds a threshold $\tau$:

\begin{equation}
\vspace{-2mm}
\text{sim}(\mu_i, \mu_j) = \frac{\mu_i \cdot \mu_j}{\|\mu_i\| \|\mu_j\|} \geq \tau
\end{equation}

Second, there must be significant overlap between the clusters. For instance, consider a counterexample where cluster $C_i$ is a large circle with 1,000 members, and $C_j$ is a single point at the center, merging would be inappropriate despite centroid proximity due to minimal overlap. Overlap is quantified as the intersection of cluster regions $\text{Area}(C_i \cap C_j)$ relative to their union $\text{Area}(C_i \cup C_j)$, satisfying:
\begin{equation}
\vspace{-2mm}
\frac{\text{Area}(C_i \cap C_j)}{\text{Area}(C_i \cup C_j)} \geq \epsilon.
\end{equation}

Both conditions ensure the merged cluster represents the data accurately without adding ambiguity or excessive variance.

\section{Results}\label{sec:result}
\subsection{Experimental Setup}
\paragraph{Dataset}
We conduct our experiments on a combination of two datasets: the Instruction-Following Eval (IFEval) benchmark \cite{zhou2023instruction} and ChatbotArena \cite{chiang2024chatbot}. IFEval consists of \textit{verifiable instructions}—prompts with programmatically checkable criteria, such as "write more than 400 words" or "mention the keyword `AI' at least three times". In contrast, ChatbotArena offers a broader set of real-world LLM use cases spanning diverse topics including language understanding, creative writing, reasoning, logic, and science. These prompts are less structured and more reflective of open-domain interactions. To balance the dataset sizes, we downsampled ChatbotArena to match the scale of IFEval.

\paragraph{Critique Model}
We use off-the-shelf language models as critique models, augmenting them with checkable rubrics to perform pairwise comparisons. Due to cost considerations, we primarily employ Gemini-1.5-Pro as the critique model. However, this choice raises the potential concern of bias toward its own outputs. To address this, we replicate the critique process using GPT-4o on a subset of 1,000 randomly selected response pairs generated by Gemini-1.5-Pro and GPT-4o. The agreement between the two critique models, measured by Pearson correlation, is 0.65—indicating a moderately strong level of reliability.

\paragraph{Hierarchical Clustering}
We perform agglomerative clustering using the cosine similarity of text embeddings derived in Section \ref{subsec:structure_critique}. To facilitate human interpretation of the resulting tree structure, we select two horizontal levels to slice the dendrogram: a fine-grained level that captures more specific behaviors (e.g., write a riddle), and a higher, more abstract level that aggregates broader categories (e.g., generate creative text). To identify the optimal clustering level, we employ the elbow method to determine the ideal number of clusters. An example of the resulting output report is shown in Figure \ref{fig:analysis_report_2_p1} and Figure \ref{fig:analysis_report_2_p2} in the appendix.

\subsection{Verifying \FrameworkName{} Reliability}
Assuming the critiques are reliable and given that the success rates are algorithmically calculated, the only potential source of error in our framework arises from the unsupervised clustering process.\footnote{\FrameworkName{} works with any critiques or rationales, whether human-provided or automatically generated. While improving LLM evaluators is beyond the scope of this paper, advancements in neural evaluators will naturally enhance the framework's reliability.} 

We judged the accuracy of our unsupervised cluster with human evaluation, as reported below:

\paragraph{Accuracy of Clustering}
We recruit human annotators to evaluate the similarity between pairs of user requests sampled from atomic claims (\textit{e.g.,} <writing lyrics that are less cliché, calculating RAM occupation>), rating similarity on a scale from 1 (completely different) to 5 (highly similar). Each input is rated by 3 annotators, resulting in 1,590 annotations. Detailed guidelines are shown in Figures \ref{fig:interface_p1} and \ref{fig:interface_p2} in Appendix \ref{appendix:annotation_guideline}. Comparing these human annotations with our embedding similarity produces a Pearson correlation of 0.643 (\textit{p}<0.0001), indicating substantial agreement.

Next, we converted the human-provided 5-point scale scores into binary categories, where scores of 4 and 5 indicate the same cluster and 1 and 2 indicate the opposite. Ambiguous pairs with a score of 3 were excluded, leaving 993 cases. These remaining cases were then divided into validation (for optimal slicing threshold) and test sets (for evaluation). Table \ref{table:clustering_success} shows that hierarchical clustering achieved a true positive rate of 0.916 and a true negative rate of 0.88. 

\begin{table}[]
\small
\centering
\begin{tabular}{@{}lll@{}}
\toprule
                & Predicted Neg. & Predicted Pos. \\ \midrule
Actual Neg. & TN = 0.883         & FP = 0.084         \\
Actual Pos. & FN = 0.117         & TP = 0.916         \\ \bottomrule
\end{tabular}
\caption{The performance of our clustering algorithm.}
\vspace{-1em}
\label{table:clustering_success}
\end{table}

\paragraph{Accuracy of Anchoring}

We evaluated our anchoring procedure using dendrograms from three independently derived model sets (Llama3, Gemini1.5, Claude3). Slicing these dendrograms at the same threshold yielded 54, 58, and 55 clusters, respectively. Human annotators reviewed 30 random members from each cluster to decide on merging, establishing a gold standard. These clusters were split into validation and test sets. We apply grid search to optimize thresholds $\tau$ and $\epsilon$ on the validation set. Test results show our merging algorithm achieved a precision of 0.926 and a recall of 0.980. 

In summary, both confirm the effectiveness of our approach in constructing and merging clusters.

\begin{figure}[t!]
    \centering
    \begin{subfigure}[t]{\linewidth}
        \centering
        \includegraphics[width=\linewidth]{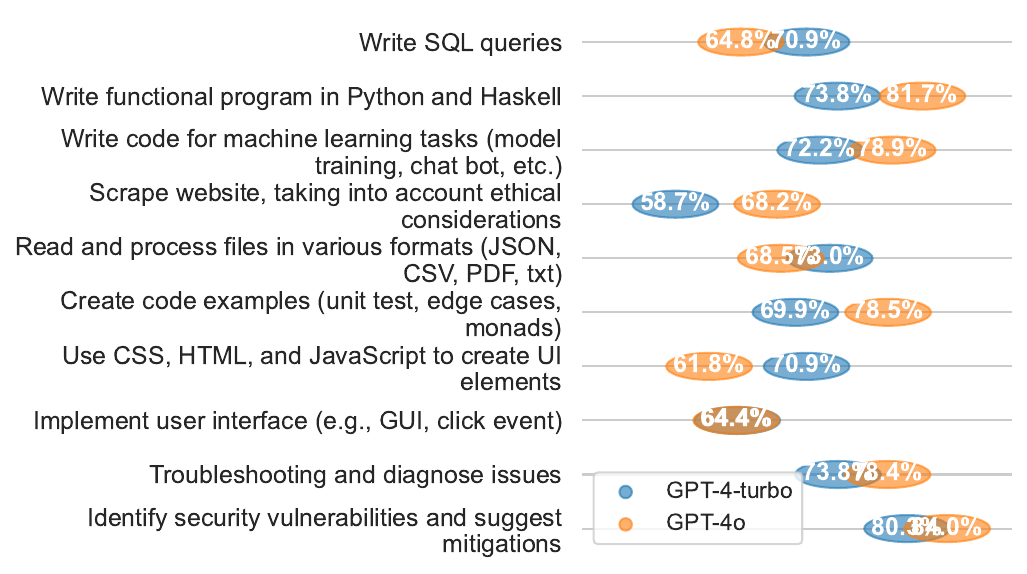}
        \caption{Performance of GPT-4-turbo and GPT-4o on coding.}
    \end{subfigure}
    \begin{subfigure}[t]{\linewidth}
        \centering
        \includegraphics[width=\linewidth]{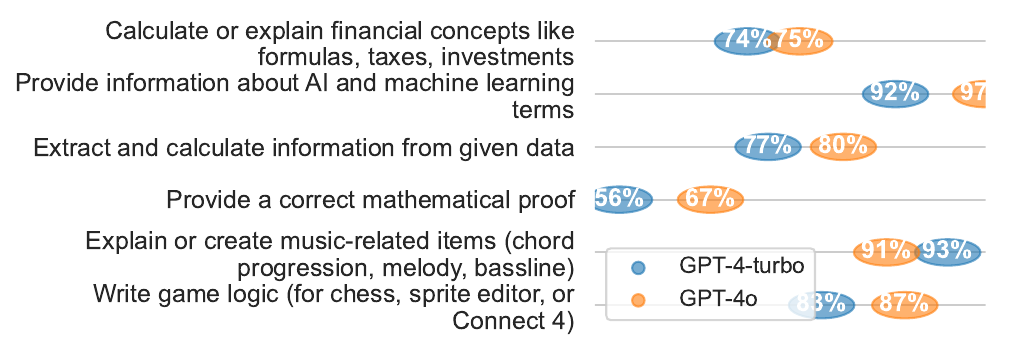}
        \caption{Results on other STEM areas.}
    \end{subfigure}
    \vspace{-2mm}
    \caption{Comparing the strengths and weaknesses of two proprietary models that were release consecutively.}
    \label{fig:comparison_gpt4o_turbo}
    \vspace{-2mm}
\end{figure}

\subsection{Insights of Fine-Grained Model Behavior}
While our framework is generic and can be applied to any \{prompt, response\} pairs, we showcase a few exemplars: 1) comparison of two proprietary models that were release consecutively by the same company(\textit{i.e.,} GPT-4o vs GPT-4-turbo), 2) comparisons across different model families, and 3) comparison between large and small models within the same family.

\paragraph{Comparison within the same model family}
Figure \ref{fig:comparison_gpt4o_turbo} illustrates the performance comparison between GPT-4o and GPT-4-turbo on STEM tasks. Despite GPT-4o being a more recent and ostensibly stronger release, \FrameworkName{} reveals that GPT-4-turbo outperforms GPT-4o in specific areas, including writing SQL queries (6.1\% improvement), reading and processing files (9.1\% improvement), and handling music-related tasks (2\% improvement).

\paragraph{Comparison across different families}
Similarly, we compare the best-performing models (as of November 1, 2024) across three families: Claude 3.5-Sonnet, Gemini-1.5-pro, and GPT-4o. \FrameworkName{} reveals that \uline{Claude excels in coding and analytical tasks} such as visualization (\textit{e.g.,} 85.5\% vs 76.8\%-79.5\%), creating or using AI models, handling edge cases, and writing shell commands; \uline{Gemini performs best in developing contents} for educational purposes, game creation, and text formatting; while \uline{GPT-4o} is superior at producing mathematical proofs, and it \uline{is exceptional at inferring the user’s precise intent from vague instructions} (83.7\% vs 63.2\%). We provide a comprehensive view of the wins and losses of these models in Figures \ref{fig:analysis_report_2_p1} and \ref{fig:analysis_report_2_p2} in the Appendix.

\begin{table}[]
\small
\centering
\begin{tabular}{@{}l@{}}
\toprule
\textbf{Identified Capabilities that Follow Inverse Scaling} \\ \midrule
Wrap the entire response in double quotes           \\
Format text using markdown                          \\
Output   in JSON format                             \\
End the response with a specific phrase             \\
Following   format of limericks and rhyme scheme    \\
Include/exclude   specific phrases                  \\
Comply with the word count requirement              \\ \bottomrule
\end{tabular}
\caption{\FrameworkName{} identified capabilities that follow inverse scaling, where increasing model size deteriorates performance.}
\label{tab:inverse_scaling_list}
\vspace{-1em}
\end{table}

\paragraph{Are larger models always better than its smaller counterparts?}
Another interesting finding emerges when comparing large and small models within the same family: Gemini-1.5 (Pro vs. Flash), Llama3.1 (405B, 70B, and 8B), and Claude3 (Opus, Sonnet, Haiku). On average, larger models outperform smaller ones across over 95\% of identified capabilities, including STEM, problem-solving, and writing tasks. However, there are a few exceptions that demonstrate \uline{inverse scaling}, where increasing model size deteriorates performance \cite{mckenzieinverse}. \FrameworkName{} discovers inverse scaling on tasks with fine-grained constraints, such as keyword inclusion/exclusion and strict formatting, as detailed in Table \ref{tab:inverse_scaling_list}.



\section{\FrameworkName{} Enhances Model Performance at Inference Time}\label{sec:ICL}
The remainder of this paper explores two extended tasks.
In this section, we illustrate how \FrameworkName{} serves as a knowledge base of model proficiency and helps improve inference-time performance by providing better few-shot demonstrations that considers both relevance and challenges posed to the target language model. In Section \ref{sec:extrapolate}, we demonstrate how the the uncovered model proficiency can serve as a foundation to predict model failures in previously unseen scenarios.

\begin{figure}[t!]
    \centering
    \includegraphics[width=0.99\linewidth]{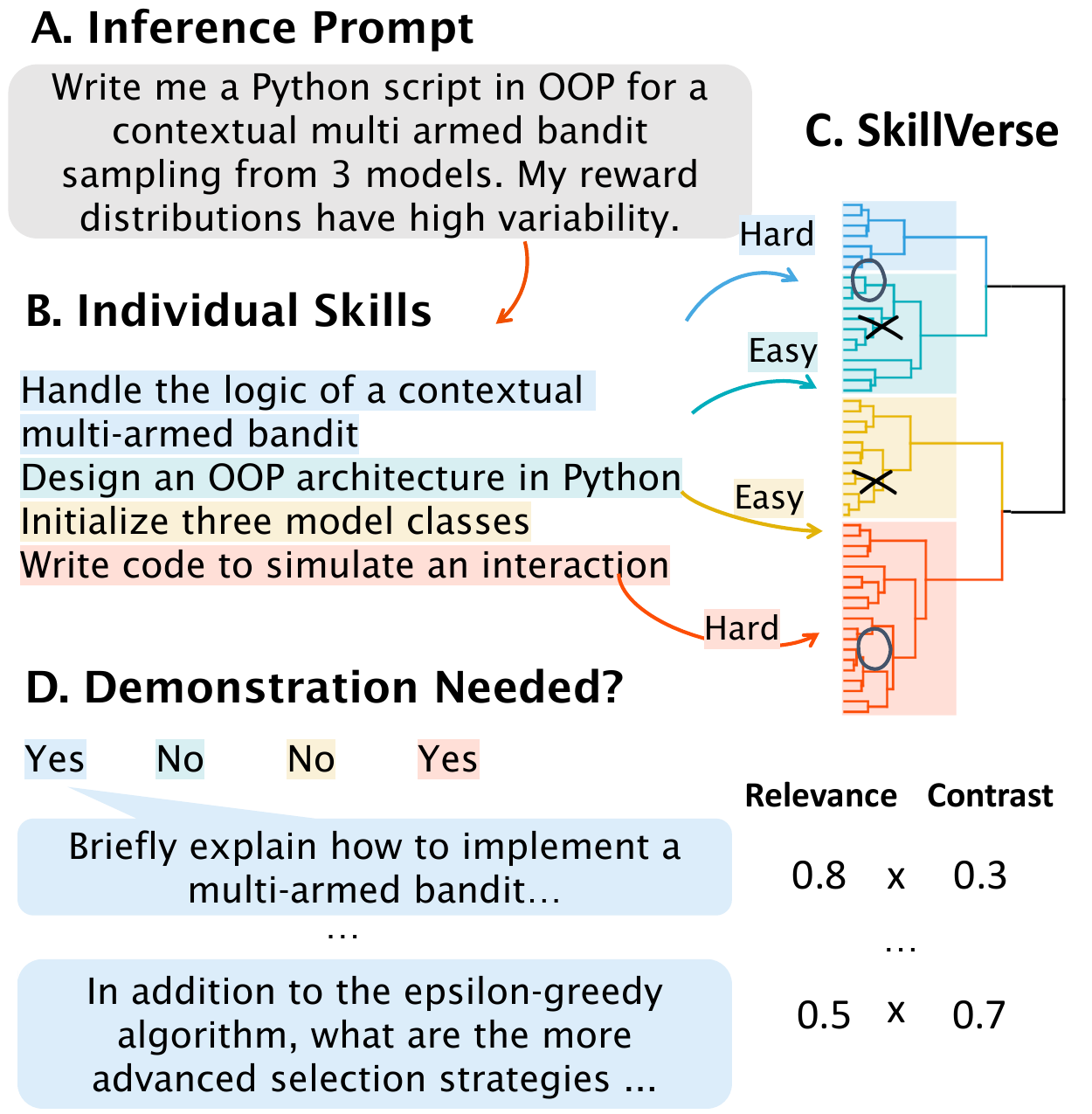}
    \vspace{-4mm}
    \caption{The dendrogram produced by \FrameworkName{} helps to selects more informative few-shot demonstrations by considering both relevance and challenges posed to the target model. In contrast, previous methods selected \textit{`...output a Python script in OOP for a bandit-inspired approach to optimize hyper-parameters across 3 models.'} which is semantically most similar but helped less as an in-context example. 
}
    \label{fig:ICL}
    \vspace{-4mm}

\end{figure}

\subsection{Approach}
\paragraph{Motivation}
Contrastive in-context learning (C-ICL), which presents an LLM with both correct and incorrect examples as demonstration, have been shown to effectively guide the models in distinguishing between desired and undesired outputs across various tasks such as information extraction \cite{chao2024context} and reasoning \cite{chia2023contrastive,zhang2024context}.

However, a typical method to construct contrastive examples involves synthetically generating negative responses by introducing hand-crafted error types, which may not best reflect a model's own distribution. Moreover, errors in C-ICL may arise from models ``over-reflecting'' on simple prompts that LLMs already know how to answer. We hypothesize that \FrameworkName{} mitigates the first issue by naturally storing pairs of good and bad responses, thereby facilitating an LLM's ability to learn from its own mistakes. Additionally, access to detailed model proficiency helps resolve the second issue, as it allows us to dynamically determine whether---and which part of---an inference prompt poses more challenge to the target model.

\paragraph{Method} 
Figure \ref{fig:ICL} illustrates the three steps to select few-shot examples with \FrameworkName{}:

\uline{Step 1: Skill Identification.} Given an inference prompt, an LLM analyzes and predicts the individual skills required to solve the task

\uline{Step 2: Mapping and Pruning.} The identified skills are located within an existing dendrogram, where simpler branches (\textit{e.g.}, those with a success rate $\geq T$) are pruned.

\uline{Step 3: Selecting Few-Shot Demonstrations.} The remaining candidate pairs are re-ranked based on two factors: (1) content relevance and (2) the benefit provided by the current contrastive pair. Here, the benefit is defined as $C(r_1)-C(r_2)$, where $C(\cdot)$ denotes the scaler score labeled by the critique model, $r_1$ refers to the correct response generated by another model and $r_2$ refers to incorrect response generated by the target model.

\subsection{Experiments}

\paragraph{Compared models}
We compare with \uline{principle learning from mistakes} \cite{zhang2024context}, which prompts the model to learn from the distilled principles derived from self-made mistakes.
We also ran two ablations for selecting few-shot examples:
\uline{similarity-only} that selects semantically similar instances \cite{mo2024c}, with or without incorporating self-generated errors as negative responses. 

\paragraph{Data Used in In-Context Learning}
We evaluated our approach to select few-shot demonstrations on GPT-4o, Gemini-1.5-pro, and Gemini-1.5-flash using two datasets: a well-structured IF-Eval dataset that involves instruction following such as format \cite{zhou2023instruction}, and the less structured ChatbotArena that involves reasoning tasks \cite{chiang2024chatbot}. We conduct the diagnosis process of \FrameworkName{} using 450 prompts from the first dataset and 2,500 prompts from the second, with inference performed on 150 held-out prompts for each dataset. Additional details about the experimental setup can be found in $\S$
\ref{appendix:extrapolation}.

\paragraph{Results}
\begin{figure}[t!]
    \centering
    \includegraphics[width=0.99\linewidth]{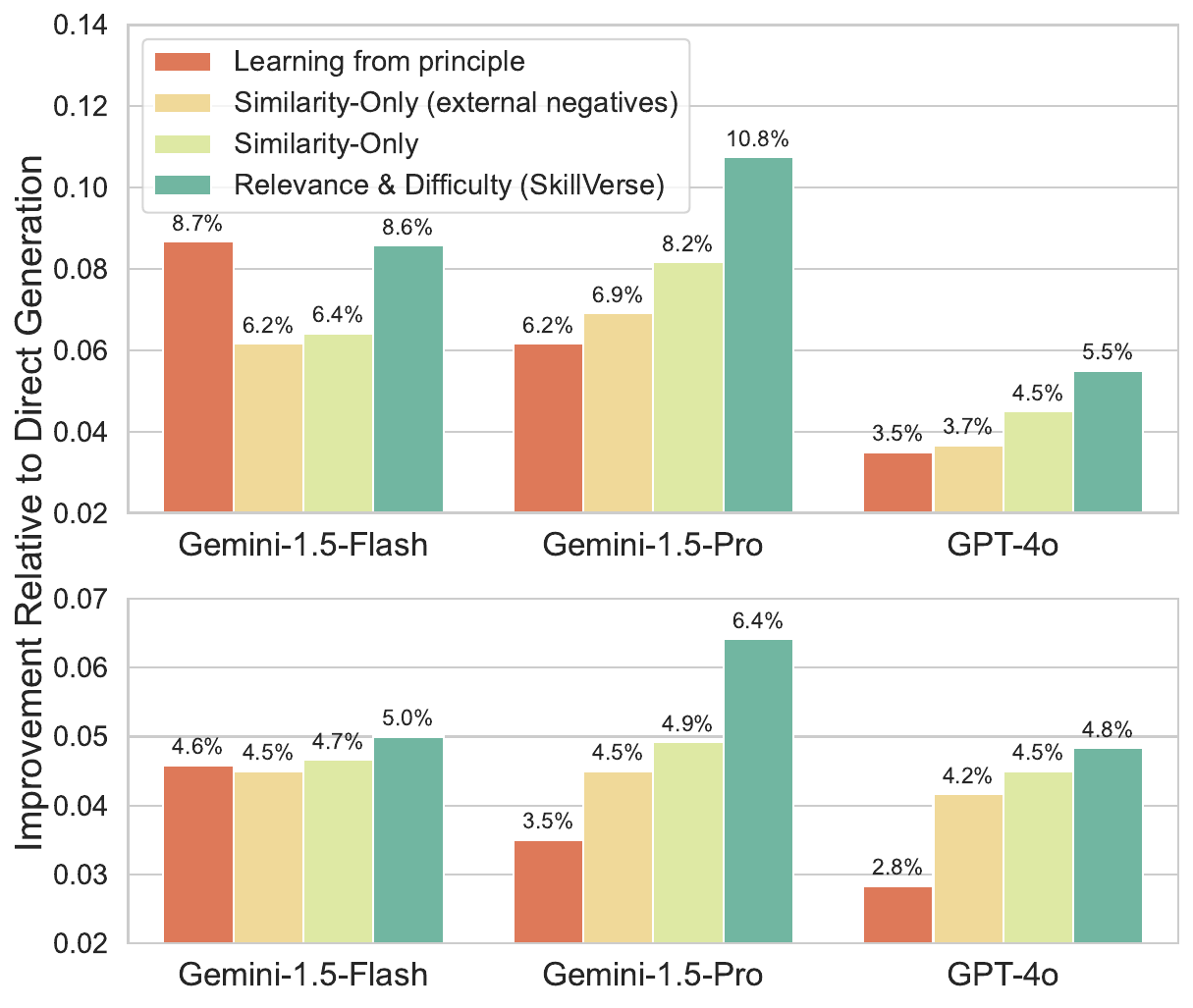}
    \vspace{-2em}
    \caption{Improvement of different in-context learning approaches, compared to direct generation on IF Eval (\textbf{top}) and Chatbot Arena (\textbf{bottom}). We posit that the performance gain is smaller for GPT-4o because of the strong performance of direct generation.}
    \vspace{-1em}
    \label{fig:ICL_preformance}

\end{figure}

We present the performance under different ICL settings in Figure \ref{fig:ICL}. 
Interestingly, we find that `learning from principles' works well with smaller models such as Gemini-1.5-flash. One possible explanation is that smaller models have limited capacity for reasoning about correct and incorrect answers in long contexts. Therefore, directly providing high-level principles might be a more effective strategy. Overall, \FrameworkName{} consistently outperforms or performs on par with all baseline models. This indicates that it successfully serves as a knowledge base of granular model proficiencies, enabling the selection of more informative in-context examples to guide the target model.

\section{Auto-Discovery: Extrapolating Model Weakness to Unseen Scenarios} \label{sec:extrapolate}

\subsection{Approach}
\begin{figure}[t!]
    \centering
    \includegraphics[width=0.95\linewidth]{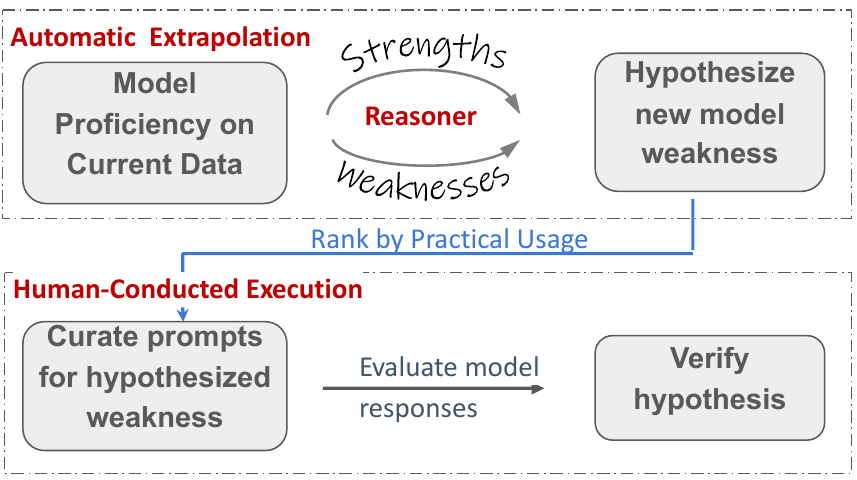}
    \vspace{-1em}
    \caption{The process to extrapolate to model deficiencies in unseen scenarios. An LLM acts as a reasoning model to generate the hypothesis, and humans execute experiments to verify these AI-generated hypotheses. 
}\vspace{-2mm}
    \label{fig:extrapolate-experiment}
\end{figure}

 We explore the feasibility of automatically extrapolating to unseen error types where models may underperform. As is shown in Figure \ref{fig:extrapolate-experiment}, we first provide the target model's capabilities on existing data and to a reasoning LLM\footnote{A different reasoning LLM is deliberately chosen to minimize inherent biases in the target model. Specifically, we use GPT-4o as it had the strongest reasoning capabilities when our experiments were conducted.} to uncover the underlying connections between areas where models perform well and poorly. Based on this analysis, we ask the reasoning model to hypothesize potential deficiencies of the target model, based on which humans curate prompts to test these hypothesized weaknesses individually. 

\begin{figure}[t!]
    \centering
    \includegraphics[width=0.9\linewidth]{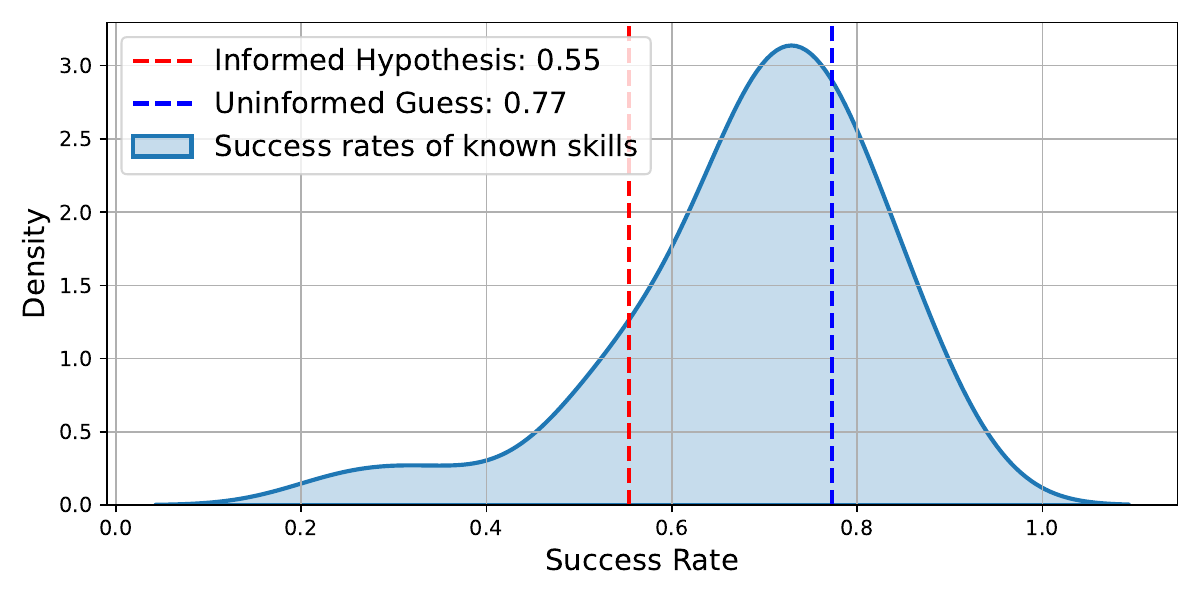}
    \vspace{-1em}
    \caption{The performance of Gemini-1.5-flash under different settings. The success rate of \FrameworkName{}-informed hypothesized weakness is only 55\%, 
    22\% lower than the uninformed hypothesis. We also list exemplar hypothesized weakness and the corresponding model performance in Table \ref{tab:extrapolation_task_1}. 
}\vspace{-1em}
    \label{fig:success_rate_plot}
\end{figure}

\begin{table*}[t!]
\small
\centering
\begin{tabular}{p{0.03\textwidth}|p{0.35\textwidth}p{0.05\textwidth}|p{0.35\textwidth}p{0.05\textwidth}}
\toprule
\rowcolor[HTML]{C6E0B4} 
\textbf{ID} & \textbf{\FrameworkName{}-Informed Hypothesis} & \textbf{Succ.} & \textbf{Uninformed Hypothesis} & \textbf{Succ.} \\ \midrule

\rowcolor[HTML]{E2EFDA} 
1 &
  Multilingual Code Switching: Seamlessly alternating between two or more languages. &
  0.607 &
  Opposing Opinions: Present two opposing opinions with equal depth and justification. &
  0.982 \\

2 &
  Logical Relations: Avoid or include words based on logical rules (e.g., AND, XOR, conditional). &
  0.148 &
  Encode Hidden Information: Insert a hidden message using techniques like acrostics or word placements. &
  0.622 \\
\rowcolor[HTML]{E2EFDA} 
3 &
  Avoid Specific Phonemes: Write text excluding words with selected phonemes (e.g., "th"). &
  0.271 &
  Physical Uncommonsense: Write a story where physical laws are broken (e.g., objects floating, time moving backward). &
  1.00 \\
4 &
  Argument Construction: Develop a three-part argument (premise, reasoning, conclusion). &
  0.506 &
Speech Impediments: Create dialogue with a specific impediment or linguistic quirk. &
  0.704 \\
\rowcolor[HTML]{E2EFDA} 
5 &
  Dynamic Math Puzzles: Create riddles where each solution depends on the previous one. &
  0.460 &
  Rhyming with Meaning: Write a poem where rhyming words form a meaningful phrase. &
  0.966 \\
\bottomrule
\end{tabular}
\vspace{-2mm}
\caption{Comparison of \FrameworkName{}-informed and uninformed predicted model weaknesses, with Gemini-1.5-flash success rates. The Kolmogorov-Smirnov test confirms statistically different distributions (p-value = 0.02).
}
\vspace{-1em}
\label{tab:extrapolation_task_1}
\end{table*}

\subsection{Experimental Setup}

We conduct experiments under \textbf{two settings}: \uline{1) identify new weaknesses in a single model}, \textit{e.g.,} Gemini-1.5-flash, and \uline{2) predict inverse scaling},, where a larger model underperforms its smaller counterpart. \textit{e.g.,} Claude-3-Opus underperforming Claude-3-Sonnet on certain skills. We ask the reasoning model to hypothesize 50 tasks for the first setting and 20 for the second. To filter out less significant predicted tasks, such as ``writing a paragraph in alternating capital and small letters'', we re-ranked them by practical relevance and selected the top half. For each selected task, we then collected 150 user prompts to test model capability solely on this task, gathered model responses, and evaluated model success rates.

As \textbf{an uninformed baseline}, we test the reasoning model’s ability to predict weaknesses without performance data. Specifically, we give it a random subset of skills and prompt it to propose new tasks where the model may fail. Comparing these uninformed predictions to informed ones reveals whether meaningful extrapolations arise from the reasoning model inherently or from \FrameworkName{}.

\begin{figure}[t!]
    \centering
    \includegraphics[width=0.99\linewidth]{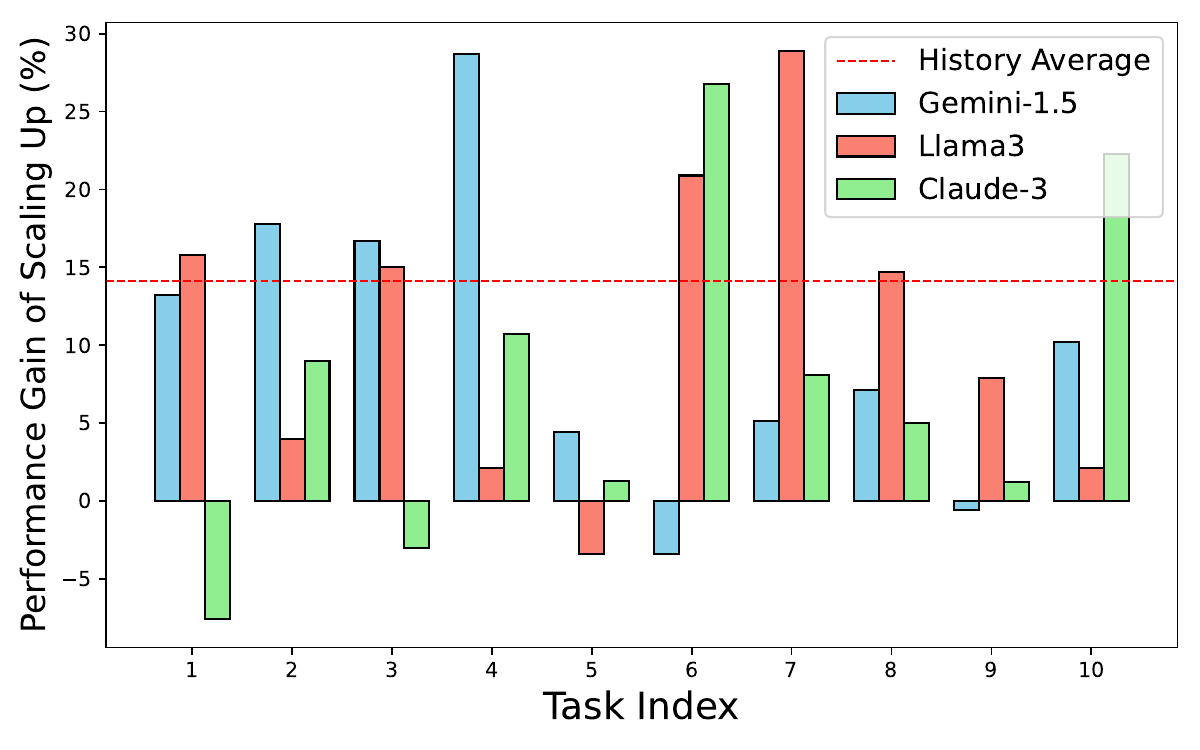}
    \vspace{-2em}
    \caption{Ten hypothesized tasks that are likely to follow inverse scaling, and the actual performance gain of scaling up. These hypotheses include \textit{formatting a bibliography in APA style}, and \textit{inserting hyperlinks into a document}, as listed in Table \ref{tab:extrapolate_task2}. Positive values show larger models outperforming smaller ones; negative values show underperformance. We show the average gain from existing data as a horizontal reference line. 
}\vspace{-1em}
    \label{fig:hypothesized_task_bars}

\end{figure}

\subsection{Result}

We visualize Gemini-1.5-flash's success rate on \FrameworkName{}-informed predicted weaknesses in Figure \ref{fig:success_rate_plot}. These predicted tasks are, on average, 14.2\% more challenging than existing tasks and 22\% more challenging than uninformed predictions. As shown in Table \ref{tab:extrapolation_task_1}, the reasoning model successfully predicted weaknesses in following logical relations (14.8\%) and avoiding specific phonemes (27.1\%). In contrast, without the contrastive insights provided by \FrameworkName{}, the same reasoner wrongly predicted weaknesses in presenting opposing opinions (98.2\%) and contradicting physical realities (100\%)--—tasks where the model actually excelled.

Moreover, as is shown in Figure \ref{fig:hypothesized_task_bars}, the reasoner also succeeds in identifying capabilities where stronger models may underperform their weaker counterparts. Appendix \ref{appendix:extrapolation} lists the predicted tasks for inverse scaling, both with (Table \ref{tab:extrapolate_task2}) and without (Table \ref{tab:extrapolation_appendix_2}) the findings by \FrameworkName{}. Under the informed setting, the average performance gain of scaling up is merely 0.5\%, which is statistically different from the 10.6\% gain observed in uninformed predictions. Both results highlight the value of \FrameworkName{} in predicting unseen model weaknesses, enabling proactive identification of potential limitations before deployment, rather than merely fixing issues after they arise.

\section{Related works}
\paragraph{Interpreting Model Behaviors.} Shifting focus from aggregated leaderboard metrics, researchers have been striving to interpret model losses more effectively. For instance, LLMSys~\cite{chiang2024chatbot}  uses BERTopic to embed prompts, reduce dimensionality, and cluster them into a predefined number of groups. QualEval~\cite{murahari2023qualeval} and a concurrent work, SkillIndex \cite{moayeri2024unearthing}, identify attributes like subtasks and domains from evaluation data and then assign them to individual data points. In contrast, \FrameworkName derives hierarchical clusters entirely unsupervised. Its tree structure enables efficient tracing of semantically similar prompts for downstream tasks while giving users control over granularity—chopping the tree at lower levels provides finer-grained loss categories for model capabilities.

 \paragraph{LLM as Evaluator.} Recently, LLM-based evaluation \cite{alpaca_eval} that requires the LLM to provide critiques to responses across a wide range of domains have emerged a scalable method for approximating human preferences \cite{zheng2023judging,chang2024survey}.
\citet{vu2024foundational} and \citet{wang2023shepherd,wang2024self} have demonstrated that critique models improve agreement with human judgment and reduce bias of the assessments when supervised multi-task fine-tuning is used. As a result, LLM-as-a-judge offers a practical alternative to traditional, labor-intensive methods of human preference collection and reward modeling \cite{wang2023shepherd, yuan2024self}.

\paragraph{Learning from Mistakes.}  
With discovered losses, how can we further improve the model? Extensive research has explored both training-time correction and inference-time improvement from mistakes or feedback~\cite{pan-etal-2024-automatically} (LLMRefine~\cite{xu-etal-2024-llmrefine}, SelfRefine~\cite{selfrefine}, etc.). \FrameworkName enhances model performance during inference. Among prior works, contrastive chain-of-thought prompting~\cite{chia2023contrastivechainofthoughtprompting} and principle learning from mistakes~\cite{zhang2024context} are most relevant as both leverage model mistakes via in-context learning. However, the few-shot demonstration examples are fixed and predefined in these works, whereas \FrameworkName adaptively selects examples for in-context learning through its dendrogram, balancing semantic relevance and potential benefit.

\section{Conclusion}
We developed a hierarchical diagnosis framework that distills a tree of fine-grained model capabilities from unstructured traffic data. 
Our framework offers the following key benefits: 1) it provides flexible insights into nuanced model abilities that are not captured by existing leaderboards or benchmarks, 2) \FrameworkName{} serves as a knowledge base of model proficiency and helps enhance the model at inference-time by providing better few-shot demonstrations, 
and 3) it can be used to predict unseen error types before deployment.


\section*{Limitation}
As pointed out by \citet{murahari2023qualeval}, fine-grained model analysis does not reject the use of benchmark metrics but uses them as one of the parts of a more actionable evaluation.

One limitation of \FrameworkName is that we use LLMs as judges to generate critiques of model responses to user prompts, which might introduce errors. Although out-of-scope of this work, developing more robust and accurate automatic critique models can definitely improve the utility of \FrameworkName. Inspired by the conclusion from prior works that large language models are better at evaluating model capabilities from comparison than evaluating the single model's response in isolation~\cite{liusie-etal-2024-llm, liusie-etal-2024-efficient}, we always compare responses from a pair of models and generate critiques. However, as recent work suggested, pairwise comparisons can sometimes amplify biases present in LLM evaluators~\cite{kawabata-sugawara-2024-rationale}. Therefore, it is crucial to be aware of potential biases that may arise during the evaluation process. In addition, as a framework designed to systematically assess model capabilities and enhance performance, we emphasize the importance of preventing the misuse of \FrameworkName.

\section*{Acknowledgements}
We would like to thank Le Hou, I-Hung Hsu, other members at Google Cloud AI, and the anonymous reviewers for the helpful discussions. 


\bibliography{custom}

\appendix

\newpage
\section*{Appendix}

\section{Full Analysis Result} \label{sec:appendix_full_result}

\begin{figure}[t!]
    \centering
    \includegraphics[width=1.0
\linewidth]{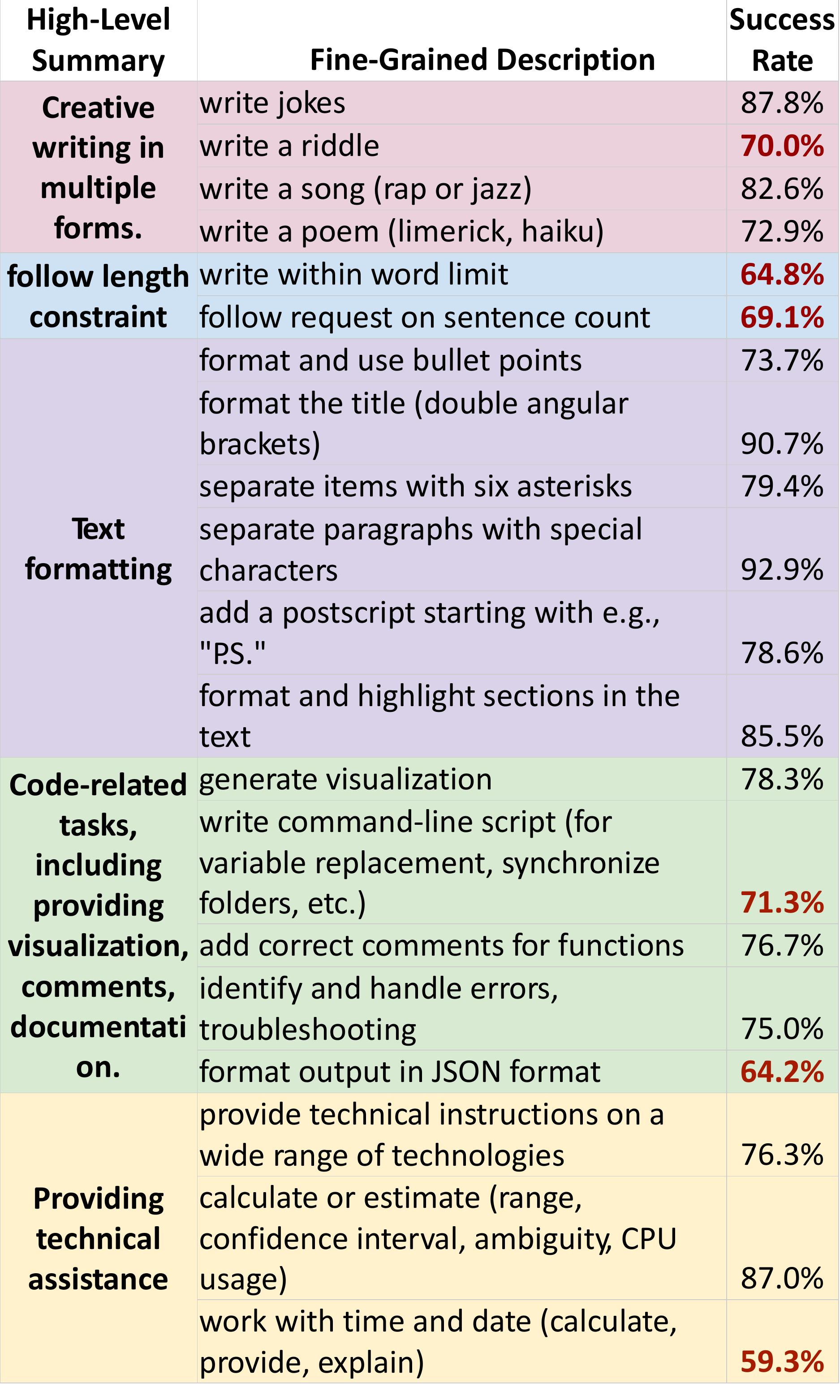}
    \caption{An example of the Gemini-1.5-flash's capability report generated by our framework. It consists of a high-level summary, descriptions of fine-grained capabilities along with the model success rates. We highlight model weaknesses in red.
    }
    \vspace{-1em}
    \label{fig:analysis_report_1}
\end{figure}

Figure \ref{fig:analysis_report_1} is an example of the Gemini-1.5-flash's capability report generated by our framework. It consists of a high-level summary, descriptions of fine-grained capabilities along with the model success rates. Overall, it is good at text formatting, and needs improvement in subdomains such as following length constraints, writing riddles, and assisting STEM tasks.

\begin{figure*}[h!]
    \centering
    \includegraphics[width=0.99
\linewidth]{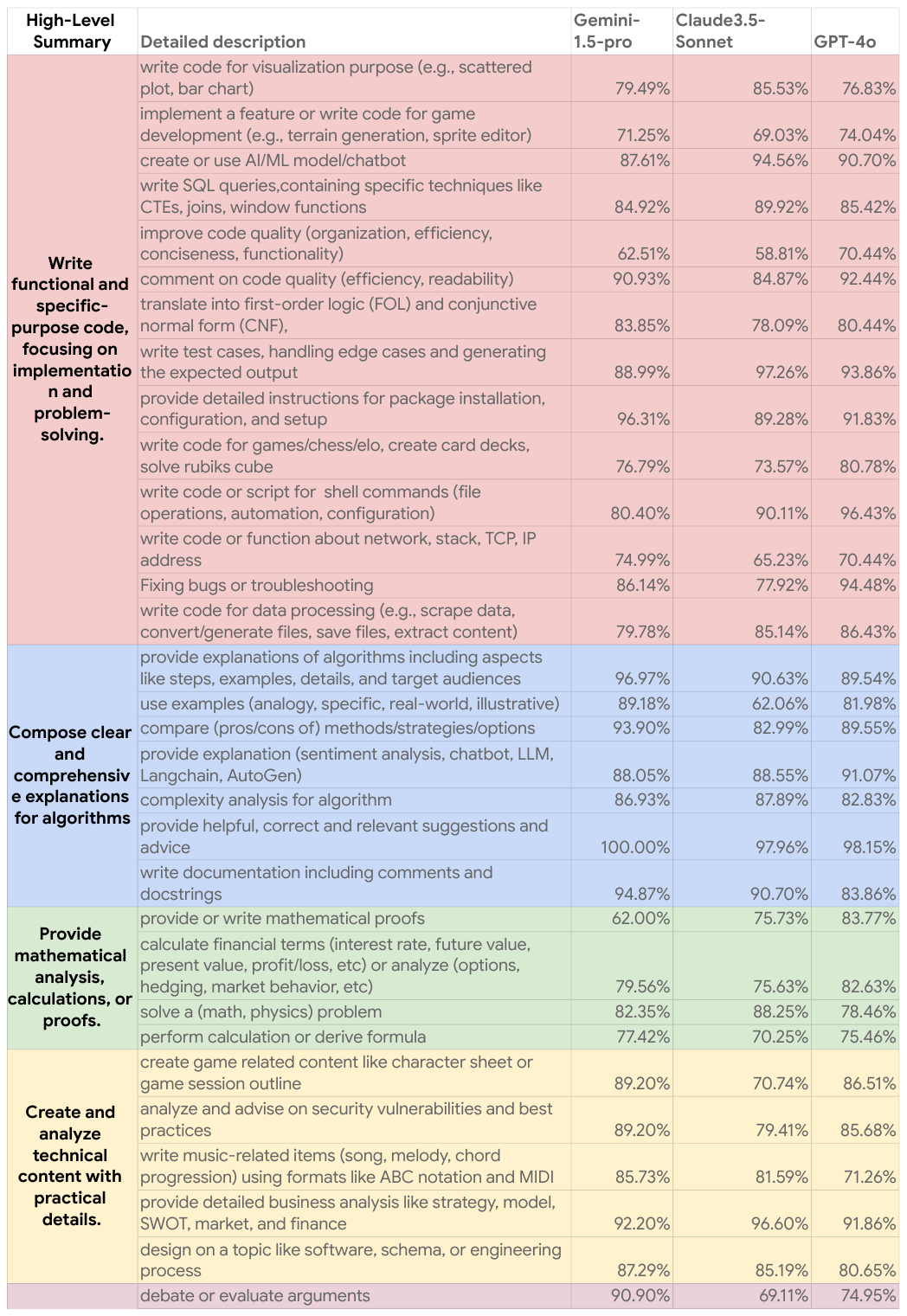}
    \vspace{-5em}
    \caption{The capability report comparing Gemini-1.5-pro, Claude3.5-Sonnet, and GPT-4o on ChatbotArena \cite{chiang2024chatbot}. It consists of a high-level summary, descriptions of fine-grained capabilities along with the model success rates.
    }
    \vspace{-1em}
    \label{fig:analysis_report_2_p1}
\end{figure*}

\begin{figure*}[t!]
    \centering
    \includegraphics[width=0.99
\linewidth]{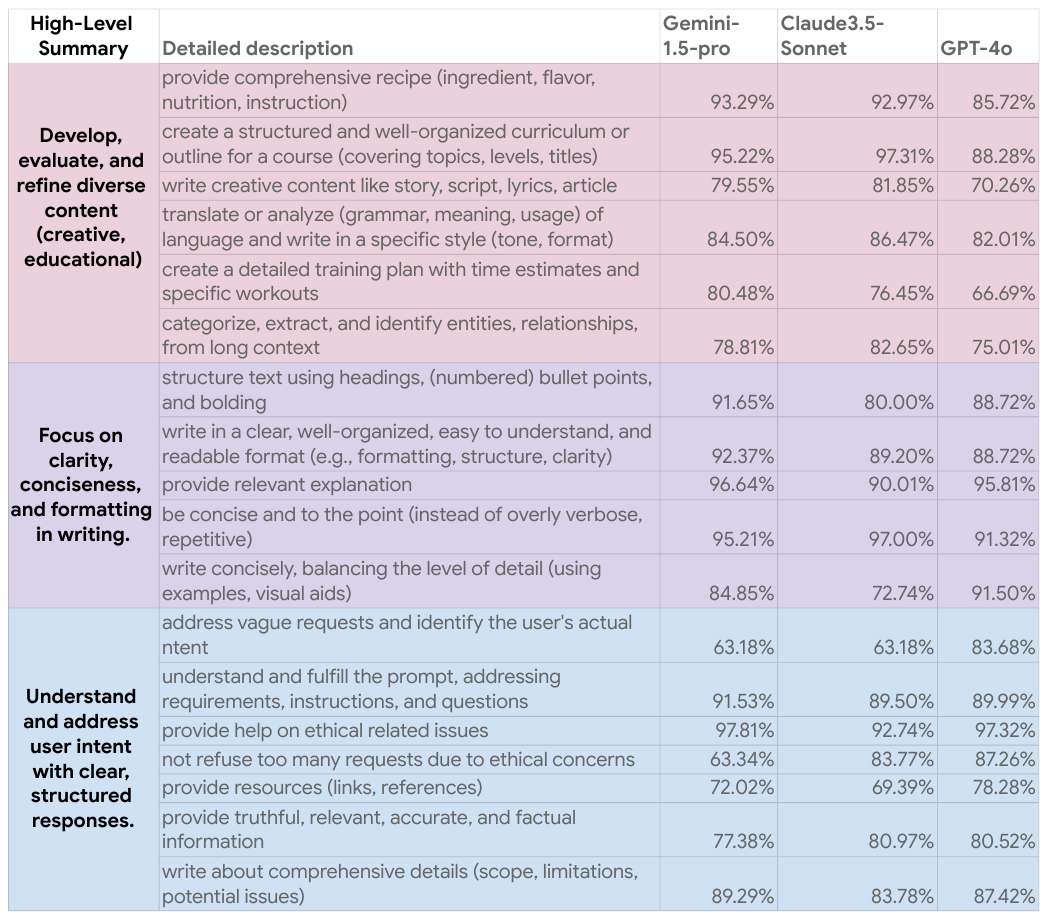}
    \vspace{-25em}
    \caption{Figure \ref{fig:analysis_report_2_p1} continued.
    }
    \vspace{-1em}
    \label{fig:analysis_report_2_p2}
\end{figure*}

Figures \ref{fig:analysis_report_2_p1} and \ref{fig:analysis_report_2_p2} present a comprehensive capability report comparing Gemini-1.5-pro, Claude3.5-Sonnet, and GPT-4o on the ChatbotArena benchmark~\cite{chiang2024chatbot}. The report includes a high-level summary, detailed descriptions of fine-grained capabilities, and the models' success rates. \FrameworkName{} reveals that Claude excels in coding and analytical tasks, such as visualization (\textit{e.g.,} 85.5\% vs. 76.8\%-79.5\%), creating or using AI models, handling edge cases, and writing shell commands. Gemini performs best in creating content for educational purposes, game development, and text formatting. Meanwhile, GPT-4o stands out in producing mathematical proofs and is exceptional at inferring user intent from vague instructions (83.7\% vs. 63.2\%).

\section{Experimental Details} \label{appendix:extrapolation}

\subsection{Detailed Results of Auto-Discovery}
Table \ref{tab:extrapolate_task2} and Table \ref{tab:extrapolation_appendix_2} lists the predicted inverse scaling tasks with or without the insights produced by \FrameworkName{}. A positive value indicates that the larger model outperforms its smaller counterpart, while a negative value indicates underperformance. On average, the larger models outperform their smaller siblings by only 0.5\%, compared to 10.6\% in the uninformed predictions.

\begin{table*}[t!]
\small
\centering

\begin{tabular}{p{0.03\textwidth}|>{\RaggedRight\arraybackslash}m{0.58\textwidth}|>{\centering\arraybackslash}m{0.06\textwidth}>{\centering\arraybackslash}m{0.06\textwidth}>{\centering\arraybackslash}m{0.06\textwidth}|>{\centering\arraybackslash}m{0.05\textwidth}}
\toprule
\rowcolor[HTML]{E2EFDA}  
\multicolumn{1}{l|}{\textbf{ID}} &
  \textbf{Description} &
  \textbf{Gemini-1.5} &
  \textbf{Llama-3} &
  \textbf{Claude-3} &
  \textbf{Avg.} \\ \midrule
\rowcolor[HTML]{FFFFFF} 
1 &
  Format a bibliography in APA style based on a list of references &
  -13.2\% &
  7.7\% &
  -13.6\% &
  -6.4\% \\
\rowcolor[HTML]{EFEFEF}  
2 &
  Insert a hyperlink for every occurrence of a specific word &
  4.7\% &
  0.4\% &
  2.3\% &
  2.5\% \\
\rowcolor[HTML]{FFFFFF} 
3 &
  Write all numbers in words instead of numerals &
  -3.4\% &
  -3.5\% &
  29.0\% &
  7.4\% \\
\rowcolor[HTML]{EFEFEF}  
4 &
  Write a response where every second sentence starts with the same word &
  -8.7\% &
  6.5\% &
  7.7\% &
  1.8\% \\
\rowcolor[HTML]{FFFFFF} 
5 &
  Use exactly one comma per sentence, placed in a specified position &
  2.9\% &
  -1.7\% &
  0.1\% &
  0.4\% \\
\rowcolor[HTML]{EFEFEF} 
6 &
  End each sentence with a specific punctuation mark (e.g., every sentence must end with an exclamation point) &
  -20.0\% &
  14.6\% &
  -6.3\% &
  -3.9\% \\
\rowcolor[HTML]{FFFFFF} 
7 &
  Format a table using LaTeX syntax &
  -9.0\% &
  5.2\% &
  5.2\% &
  0.5\% \\
\rowcolor[HTML]{EFEFEF} 
8 &
  Replace every occurrence of the word 'the' with 'a' &
  13.7\% &
  12.6\% &
  -6.5\% &
  6.6\% \\
\rowcolor[HTML]{FFFFFF} 
9 &
  Enclose the entire response in an HTML \textless{}p\textgreater{} and \textless{}/p\textgreater{} tag &
  1.8\% &
  1.8\% &
  -8.8\% &
  -1.8\% \\
\rowcolor[HTML]{EFEFEF}  
10 &
  Write a strict-structure sonnet while maintaining a particular rhyme scheme (e.g., iambic pentameter) &
  -0.2\% &
  5.3\% &
  1.6\% &
  2.2\% \\
\rowcolor[HTML]{FFFFFF} 
11 &
  Wrap conversations and individual sentences in parentheses &
  -10.9\% &
  5.9\% &
  -7.3\% &
  -4.1\% \\
\rowcolor[HTML]{CCD9D4} 
\bottomrule
\end{tabular}
\vspace{-2mm}
\caption{Predicted inverse scaling tasks (\FrameworkName{}-informed) and the performance gap between larger and smaller models. We evaluated Gemini-1.5 (pro vs. flash), Llama3 (405b vs. 70b), and Claude3 (Opus vs. Sonnet). A positive value indicates that the larger model outperforms its smaller counterpart, while a negative value indicates underperformance. On average, the larger models outperform their smaller siblings by only 0.5\%, compared to 10.6\% in the uninformed predictions (Table \ref{tab:extrapolation_appendix_2} in the Appendix). 
}
\label{tab:extrapolate_task2}
\vspace{-2mm}
\end{table*}

\begin{table*}[]
\small
\centering

\begin{tabular}{p{0.03\textwidth}|>{\RaggedRight\arraybackslash}m{0.5\textwidth}|>{\centering\arraybackslash}m{0.06\textwidth}>{\centering\arraybackslash}m{0.06\textwidth}>{\centering\arraybackslash}m{0.08\textwidth}|>{\centering\arraybackslash}m{0.06\textwidth}}
\toprule
\rowcolor[HTML]{E2EFDA}  
\multicolumn{1}{l|}{\textbf{Index}} &
  \textbf{Description} &
  \textbf{Gemini-1.5} &
  \textbf{Llama3} &
  \textbf{Claude-3} &
  \textbf{Average} \\ \midrule
\rowcolor[HTML]{FFFFFF} 
1 &
  Respond to a question using only questions &
  13.2\% &
  15.8\% &
  -7.6\% &
  7.1\% \\
\rowcolor[HTML]{EFEFEF}  
2 &
  Recommend a Playlist Based on Mood &
  17.8\% &
  4.0\% &
  9.0\% &
  10.3\% \\
\rowcolor[HTML]{FFFFFF} 
3 &
  Describe something as if it were a recipe &
  16.7\% &
  15.0\% &
  -3.0\% &
  9.6\% \\
\rowcolor[HTML]{EFEFEF}  
4 &
  Suggest a Weekly Meal Plan &
  28.7\% &
  2.1\% &
  10.7\% &
  13.8\% \\
\rowcolor[HTML]{FFFFFF} 
5 &
  Create a riddle for technical concepts &
  29.4\% &
  23.8\% &
  16.4\% &
  23.2\% \\
\rowcolor[HTML]{EFEFEF} 
6 &
  Describe a process, but only using the future tense &
  4.4\% &
  -3.4\% &
  1.3\% &
  0.8\% \\
\rowcolor[HTML]{FFFFFF} 
7 &
  Reverse Word Order &
  -3.4\% &
  20.9\% &
  26.8\% &
  14.8\% \\
\rowcolor[HTML]{EFEFEF} 
8 &
  Generate Advice for Improving Public Speaking Skills &
  5.1\% &
  28.9\% &
  8.1\% &
  14.0\% \\
\rowcolor[HTML]{FFFFFF} 
9 &
  Write a song chorus and bridge &
  7.1\% &
  14.7\% &
  5.0\% &
  8.9\% \\
\rowcolor[HTML]{EFEFEF} 
10 &
  Generate Rhyming Words &
  -0.6\% &
  7.9\% &
  1.2\% &
  2.8\% \\
\rowcolor[HTML]{FFFFFF} 
11 &
  Alliteration Generation &
  10.2\% &
  2.1\% &
  22.3\% &
  11.5\% \\

\bottomrule
\end{tabular}
\caption{Predicted inverse scaling tasks (uninformed predictions) and the performance gap between larger and smaller models. We evaluated Gemini-1.5 (pro vs. flash), Llama3 (405b vs. 70b), and Claude3 (Opus vs. Sonnet). A positive value indicates that the larger model outperforms its smaller counterpart, while a negative value indicates underperformance. On average, the larger models outperform their smaller siblings by 10.6\%, with a distribution that is statistically different from those predicted with the performance data as input.}
\label{tab:extrapolation_appendix_2}
\end{table*}

\subsection{Human Annotation}\label{appendix:annotation_guideline}
\begin{figure*}[t!]
    \centering
    \includegraphics[width=0.85
\linewidth]{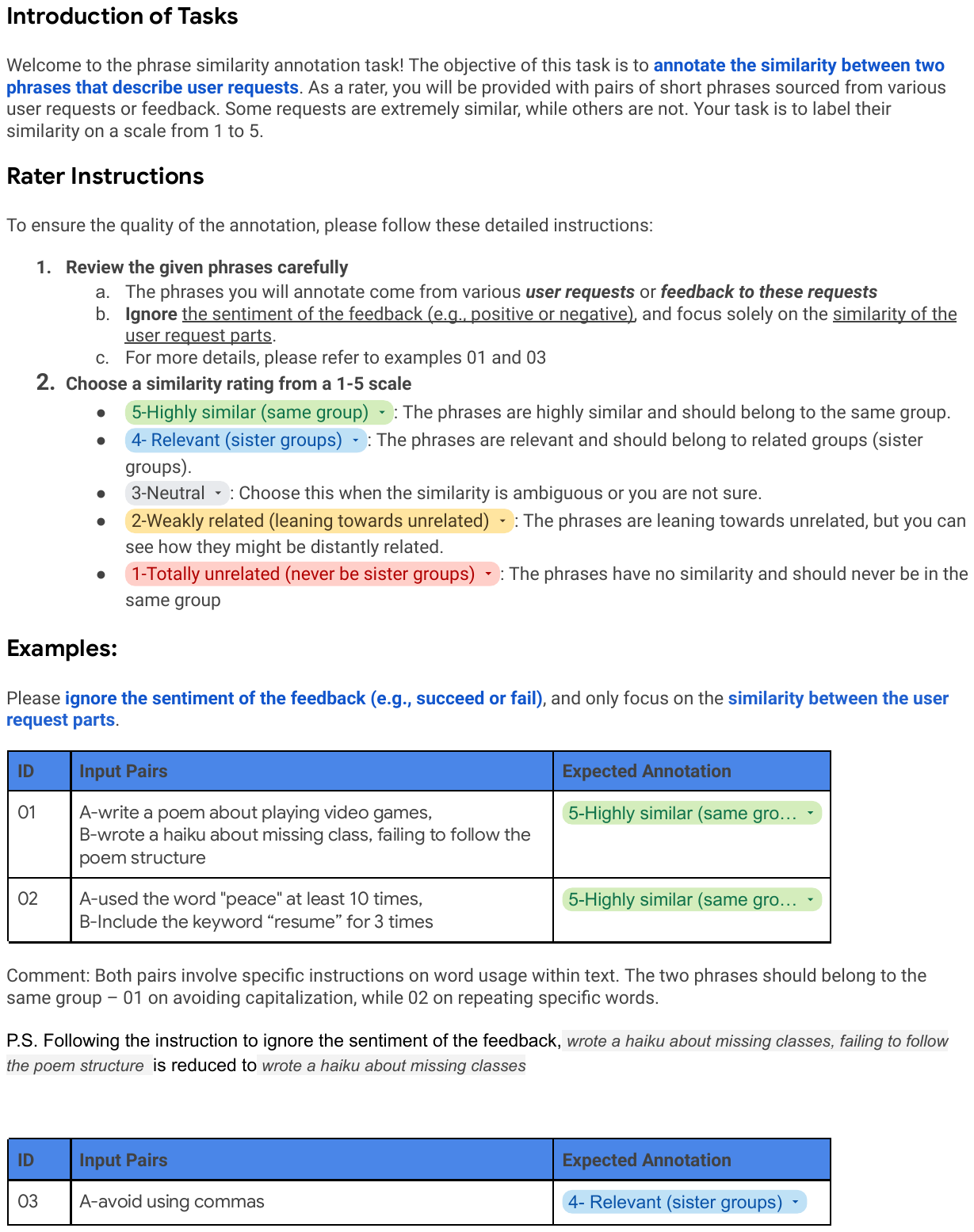}
    \vspace{-2mm}
    \caption{Annotation Guideline: Phrase Similarity Annotation Instructions Page 1}
    \label{fig:interface_p1}
   
\end{figure*}

\begin{figure*}[t!]
    \centering
    \includegraphics[width=0.85
\linewidth]{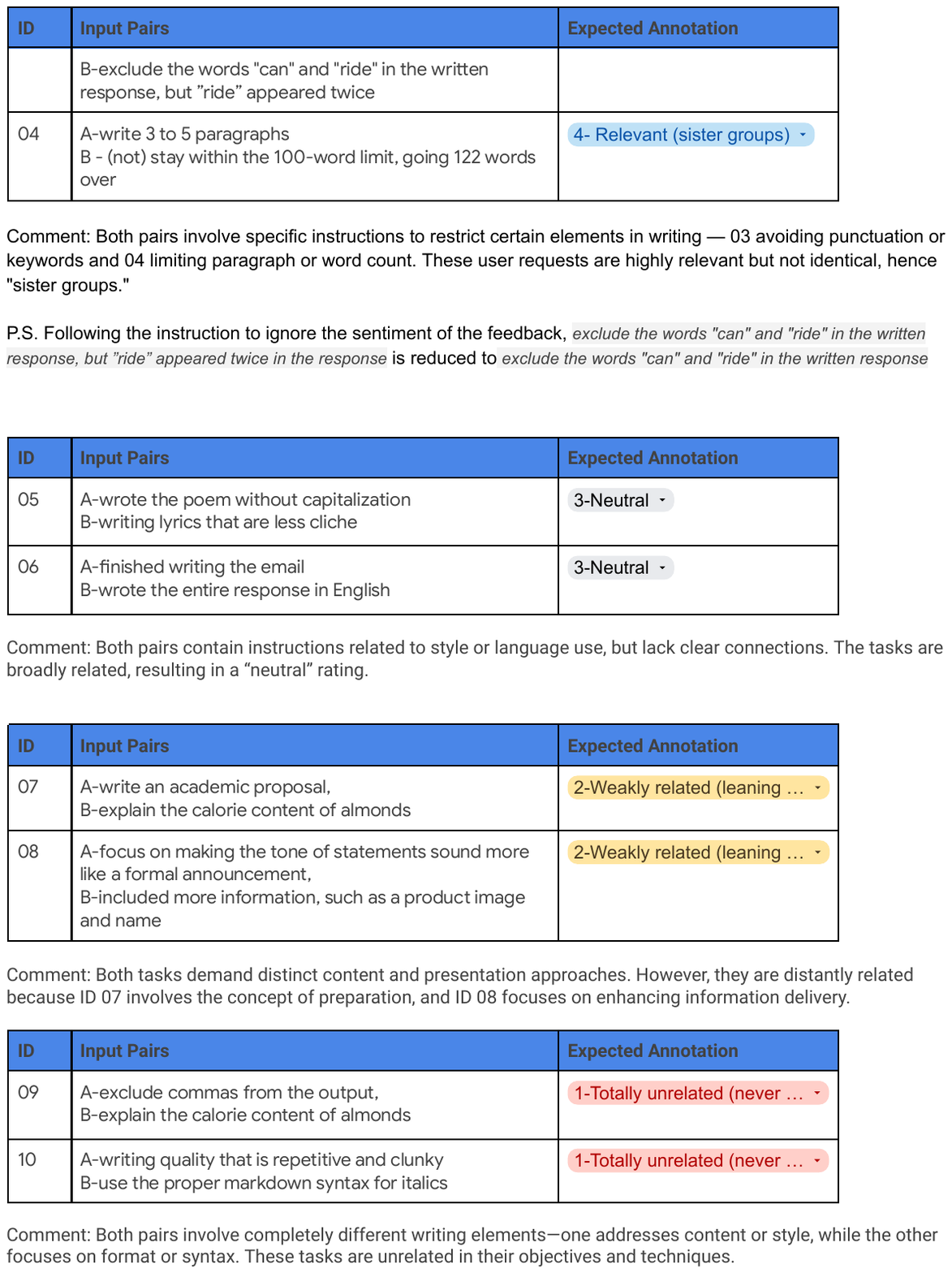}
    \vspace{-2mm}
    \caption{Annotation Guideline: Phrase Similarity Annotation Instructions Page 2}
    \label{fig:interface_p2}
   
\end{figure*}

We present the complete annotation guideline used in Section \ref{sec:result} to verify the accuracy of our clustering algorithm in Figure \ref{fig:interface_p1} and \ref{fig:interface_p2} The inter-annotator agreement is 0.88 as measured by Pearson correlation.

\end{document}